\documentclass[letterpaper, 10 pt, journal, twoside]{IEEEtran}
\usepackage[ruled,vlined,noalgohanging]{algorithm2e}
\usepackage{xcolor}
\usepackage{cite}
\usepackage{graphicx}
\graphicspath{ {images/} }
\usepackage{amsmath}
\usepackage{svg}
\usepackage{makecell}
\usepackage{subfigure}
\usepackage[cmex10]{mathtools}
\usepackage{placeins}
\usepackage{color}
\usepackage{soul}
\usepackage{float}
\usepackage{booktabs}
\usepackage{graphics}
\usepackage{enumitem}
\usepackage[normalem]{ulem}
\useunder{\uline}{\ul}{}
\usepackage{booktabs}
\usepackage[switch]{lineno}
\usepackage{amssymb}
\usepackage{siunitx}
\usepackage{multirow}
\usepackage{url}
\usepackage{url}

\usepackage{booktabs} 
\usepackage{boldline} 

\begin{document}


\title{\LARGE \bf
DBF-MA: A Differential Bayesian Filtering Planner for Multi-Agent Autonomous Racing Overtakes
}

\author{Trent Weiss$^1$, Amar Kulkarni$^2$ and Madhur Behl$^2$}
\maketitle
\begin{abstract}
A significant challenge in autonomous racing is to generate overtaking maneuvers.  Racing agents must execute these maneuvers on complex racetracks with little room for error.  Optimization techniques and graph-based methods have been proposed, but these methods often rely on oversimplified assumptions for collision-avoidance and dynamic constraints. In this work, we present an approach to trajectory synthesis based on an extension of the Differential Bayesian Filtering framework. Our approach for collision-free trajectory synthesis frames the problem as one of Bayesian Inference over the space of Composite B\'ezier Curves. Our method is derivative-free, does not require a spherical approximation of the vehicle footprint, linearization of constraints, or simplifying upper bounds on collision avoidance.  We conduct a closed-loop analysis of DBF-MA and find it successfully overtakes an opponent in 87\% of tested scenarios, outperforming existing methods in autonomous overtaking.

\end{abstract}

\section{Introduction}
\label{sec:intro}
Autonomous racing has emerged as a distinct and growing research area~\cite{betz2022autonomous}.  
Events such as the Indy Autonomous Challenge (IAC) \cite{indy}, F1tenth~\cite{babu2020f1tenth} and A2RL\cite{a2rl} routinely stage high-speed, close-proximity autonomous races where agents operate close to their dynamic limits. 
Yet recent multi-car demonstrations (e.g., Jan-2025 four-car IAC exhibition during CES) were constrained by speed caps and fixed inside/outside lines, far from the open-world, unconstrained interactions typical of human F1/Indy racing. 
Planning for high-speed, multi-agent autonomous racing under uncertainty remains open.

A central maneuver in these settings is an \textbf{autonomous overtake}. 
An effective overtake requires synthesizing a trajectory that (i) respects track limits, (ii) remains dynamically feasible for the ego vehicle, (iii) avoids any contact with opponents whose future motion is uncertain, and (iv) returns the ego to the optimal racing line (ORL) to preserve corner entry/exit speed. 
This problem is difficult for several reasons.
First, autonomous racing inherintly occurs at extreme speeds and acceleration profiles with very small error margins.
Any trajectory planning errors for the ego vehicle can have catastrophic consequences.  Higher speeds also demand computationally efficient planners. 
Second, the opponent agent's future motion is far more uncertain compared to urban/highway driving.  
Opponents are not confined to lanes or speed limits, rendering ``avoid the full reachable set" strategies impractical. 
Finally, typical overtakes are executed with only a small relative acceleration advantage, especially through corners, leaving little to no room for conservative spacing.  
Additionally, a successful overtaking maneuver should end with the ego vehicle merging back on the ORL.
This brings us to the core problem of trajectory synthesis for multi-agent autonomous racing: \textit{How can an autonomous racing agent synthesize an overtaking maneuver that stays within track limits, lies within the vehicle’s feasible control envelope, avoids collisions with uncertain opponents, and re-merges onto the ORL?}  
In our prior work, we presented a sampling-based Bayesian method - Differential Bayesian Filtering (DBF)~\cite{ral_dbf}, which addressed the problem of single-agent planning for autonomous racing. We extend the approach to multi-agent racing present the DBF-MA algorithm.
DBF-MA utilizes Bayesian Inference over the space of Composite B\'ezier Curves (CBC) to generate a distribution of trajectories that satisfies track-boundary, collision-avoidance, and dynamic feasibility constraints.  DBF-MA makes two key contributions:
\begin{enumerate}
    \item We propose a compact, $C^1$-continuous parameterization of overtaking trajectories using Composite B'ezier Curves. Key properties of CBCs guarantee satisfaction of overtake boundary conditions (start on/near ORL with an opponent ahead, execute pass, and finish back on ORL), while enabling explicit track-limit and curvature/acceleration constraints.
    \item We formulate trajectory synthesis as Sequential Monte Carlo (particle filtering) over the CBC parameter space. The likelihood encodes the probability of satisfying all the racing constraints, yielding a posterior distribution over collision-free, feasible overtaking maneuvers.
\end{enumerate}
 DBF-MA  produces risk-aware maneuvers that respect track limits, maintain feasibility, avoid collisions under opponent uncertainty, and terminate on the ORL. We evaluate DBF-MA in a simulated racing environment and show that it outperforms existing methods on several racing methods.


\begin{figure}
    \centering
    \includegraphics[width=0.85\linewidth]{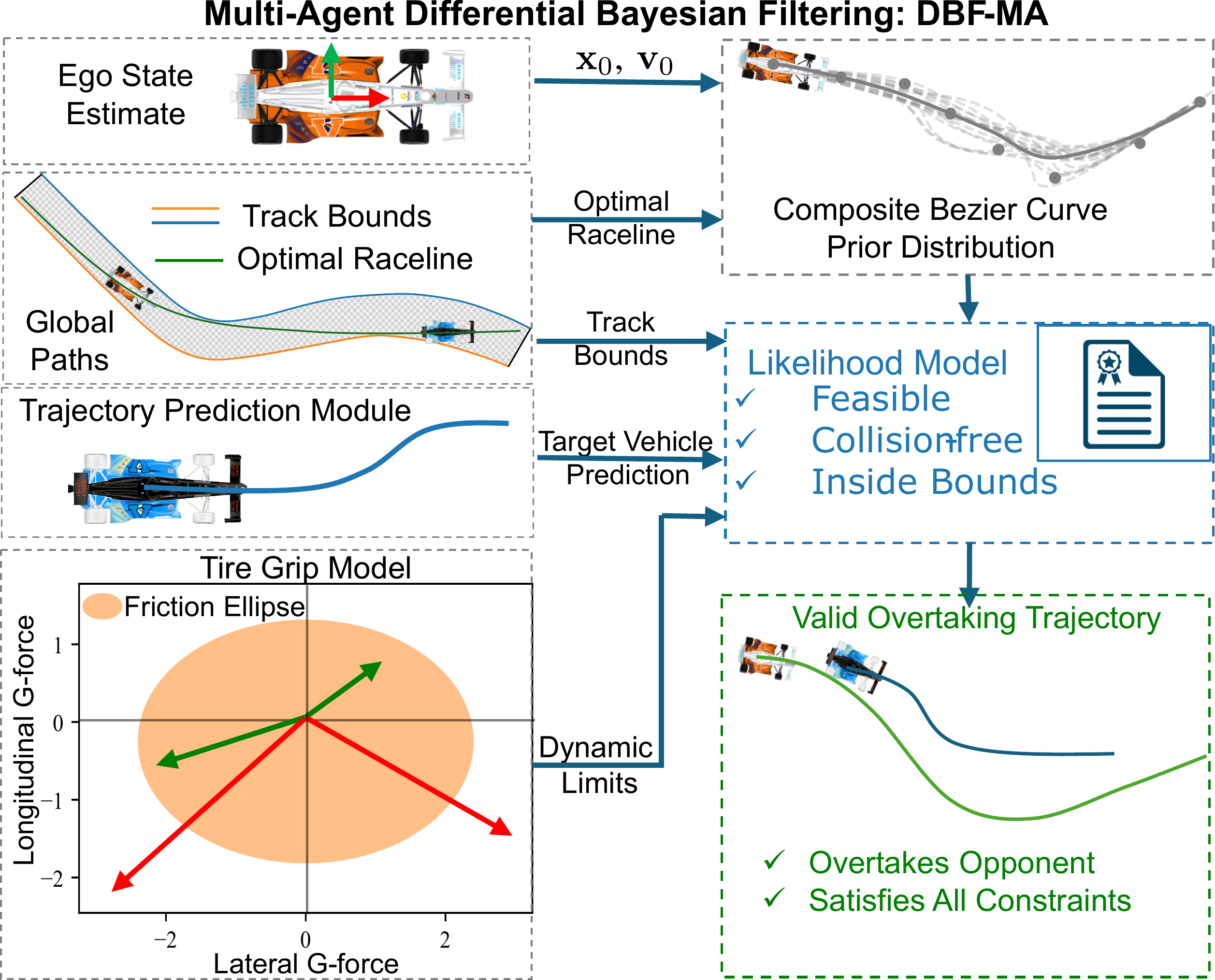}
    \caption{The Multi-Agent Differential Bayesian Filtering (DBF-MA) algorithm for overtaking trajectory synthesis.  DBF-MA frames the problem as Bayesian inference over the parameters of a Composite B\'ezier Curve.}
    \label{fig:hero_figure}
\end{figure}

\section{Related Work}
\label{sec:related_work}
Trajectory synthesis for autonomous vehicles is a well-studied topic. Model predictive control (MPC) formulations are a common approach~\cite{mpc_collisionavoidance_survery,learning_mpc,mpc_overtaking}. These works frame the problem as a constrained optimization problem to ensure desirable properties of the resultant trajectory.  However, adding collision avoidance constraints to these MPC formulations is challenging and often requires simplifying assumptions.  Many MPC-based methods model agents with a bounding sphere as its collision geometry~\cite{cvar,Frey-RSS-20,mpc_collisionavoidance_survery}. I.e. these methods impose a minimum Euclidean distance of at least 1 car diameter between vehicles as part of the optimization problem. This over-approximation of the vehicle's footprint can be justified for avoiding pedestrians, but is too conservative for racing, where wheel-to-wheel scenarios with much less than a car diameter of space between vehicles are commonplace.  The authors in \cite{gaussian_processes_mpc_overtaking} utilize a rectangular buffer zone around vehicles to overtake, but limit their analysis to highway driving and assume vehicles stay in defined lanes at prescribed speed limits, which is not applicable to racing.   Finally, many optimization methods require computing the derivative (Jacobian) of these constraint functions w.r.t. the optimization variables (control inputs), which can be difficult and/or computationally expensive for realistic racecar models. 
Our method also utilizes a rectangular buffer for collision avoidance (far more realistic for racecars, which tend to be much longer than they are wide), requires no assumptions about the structure of the opponent vehicle's predicted motion, and is built on a derivative-free method of sampling-based Bayesian inference.


Trajectory synthesis for racing has also been considered, albeit to a lesser extent.  ~\cite{graph_planner_og} and \cite{graph_planner_oval} present a graph-based approach that segments the racetrack longitudinally and orthogonal to the optimal racing line.  However, ~\cite{graph_planner_og} requires very sparse longitudinal spacing (30$m$) of the graph nodes to make the search tractable and limit their experimental analysis to overtaking scenarios with the opponent vehicle traveling at a fixed speed of 75 $kph$, which is not extensible to complex road courses where agents execute considerable acceleration and braking throughout the track. Finally, ~\cite{graph_planner_og} does not guarantee that the endpoint of the planned path will be on the optimal racing line, which could leave the ego vehicle in a state from which returning to the ORL is impossible. Our method does not have this limitation, and explicitly guarantees that the endpoint of the overtaking trajectory will be on the ORL at its prescribed velocity.


Other methods in overtaking for autonomous racing utilize Frenet coordinates to generate trajectories as offsets from a reference trajectory~\cite{og_frenet}.  ~\cite{euroracing_arch} presents a system architecture for autonomous racing with a similar approach for planning, utilizing an optimization algorithm on a Frenet space to plan trajectories. The Spliner~\cite{spliner} algorithm and it's extension, Predictive Spliner~\cite{predictive_spliner}, make heavy use of B-splines in a Frenet frame to generate collision-free overtaking trajectories, but only applies curvature limits to the synthesized plans.  These kinematic constraints suffice for Spliner's intended application on the F1/10 platform (where speeds rarely exceed 50MPH), but does not extend to scenarios where \emph{dynamic} limits are important, such as high-speed F1 or IndyCar racing. 

Reinforcement Learning (RL) has also been considered for autonomous racing \cite{rl_lee_RAL,rl_Sony}. Although RL methods demonstrate good performance, they are more susceptible to the Sim-to-Real gap.  RL techniques require thorough exploration of the entire action space and will often crash the vehicle many times during training. Given that the target application of this work is the Indy Autonomous Challenge on a full-scale real-world racecar, where deploying an RL-based technique is not feasible, we do not directly compare against RL methods.

Our method does not make the bounding-circle assumption such as in ~\cite{mpc_collisionavoidance_survery,cvar,Frey-RSS-20,learning_mpc}, does not require an expensive graph-search operation, and explicitly guarantees that the endpoint of the planned trajectory will be on the optimal racing line at it's prescribed velocity, ensuring a feasible transition back to this optimal path. We show in section \ref{sec:results} that our method also achieves better overtaking performance than both the graph based method in ~\cite{graph_planner_oval} (a sampling-based method) and Predictive Spliner (an optimization based method).
We now describe the problem of overtaking in autonomous racing more formally and how our Bayesian inference-based approach is well suited to solving it.

\section{Problem Formulation}
\label{sec:problem_formulation}
We consider an autonomous racing scenario with an ego vehicle (controlled by an autonomous agent) and a target vehicle that the ego aims to overtake. 
The problem setting involves two global elements:
\begin{enumerate}
    \item \textbf{Track boundaries.} Let $\mathbf{B}_I(s),\mathbf{B}_O(s):\mathbb{R}\to\mathbb{R}^2$ denote the inner and outer boundary curves parameterized by arc length $s$. The drivable set is the closed region $\mathbb{B}\subset\mathbb{R}^2$ between them.
    \item \textbf{Optimal Racing Line (ORL).} A $C^1$ curve denoted by $\mathrm{ORL}(s):[0,L_{ORL}]\to\mathbb{R}^2$ with position $\mathbf{p}_{ORL}(s) \in \mathbb{R}^2$ and nominal velocity $\mathbf{v}_{orl}(s) \in \mathbb{R}^2$ at each $s$. This is the known racing line around the track that may be derived using geometric methods such as min curvature, or model-based forward-backward optimization methods. \cite{raceline_generation}.
    The set of all points on the racing line is $\mathbb{ORL} = \{\mathbf{p}_{ORL}(s)\text{ }\forall\text{ }s\in [0, L_{ORL}]\}$.
    We define, $\pi_{ORL}(\mathbf{x})=\underset{s}{argmin}\text{ }\vert\vert \mathbf{p}_{ORL}(s) - \mathbf{x}\vert\vert$ as the projection of position $\mathbf{x} \in \mathbb{R}^2$ onto the ORL. $\pi_{ORL}(\mathbf{x})$ can be interpreted as the abscissa of $\mathbf{x}$ in a Frenet coordinate frame with the ORL as it's reference curve.
\end{enumerate}
\begin{figure}
    \centering
    \includegraphics[width=0.9\columnwidth]{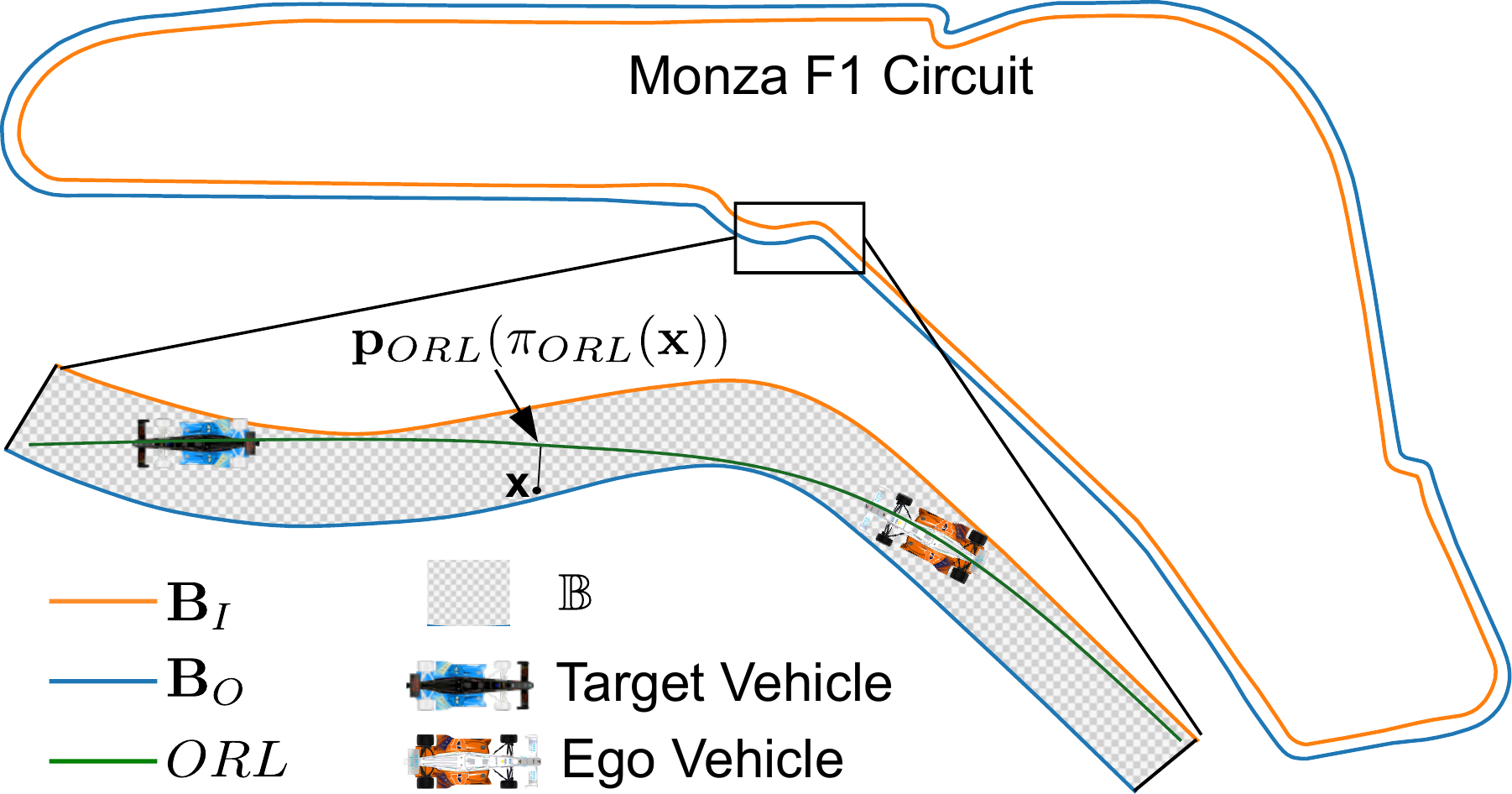}
    \caption{The local planning problem on the Monza F1 Circuit. The goal of a local racing planner is to generate a trajectory that overtakes the target vehicle and returns to the Optimal Racing Line.}
    \label{fig:vehicles_and_orl}
\end{figure}



Fig.~\ref{fig:vehicles_and_orl} shows $\mathbf{B}_I,\mathbf{B}_O,\mathbb{B}$, and the ORL for the Monza F1 Circuit, as an example. The global paths (bounds and ORL) are assumed known a priori.

The problem of autonomous overtaking can be framed as a local planner that must generate a trajectory $\mathcal{T}_{ego}(t) : [0, T_F] \rightarrow \mathbb{R}^2$, for the ego agent, given the current ego state $(\mathbf{x}_0,\mathbf{v}_0)$, the global priors, and a predicted trajectory for the target vehicle $\mathcal{T}_{target}(t)$; where $t=0$ represents the current time and $T_F$ is the planning horizon.
The ego trajectory $\mathcal{T}_{ego}(t)$, is subject to the following constraints:

\noindent\textbf{Continuity (C)}: ensures $\mathcal{T}_{ego}(t)$ starts at the current state.
\begin{equation}
\mathcal{T}_{ego}(0)=\mathbf{x}0,\quad \dot{\mathcal{T}}_{ego}(0)=\mathbf{v}0,
\tag{C}
\label{eqn:initial_pos}
\end{equation}
\noindent\textbf{Track-keeping (TK):} enforces that the ego stays within the track boundaries.
\begin{equation}
\mathcal{T}_{ego}(t)\in\mathbb{B}\quad \forall t \in [0, T_F]
\tag{TK}
\label{eqn:ontrack_constraint}
\end{equation}
\noindent\textbf{Collision Avoidance (CA):} guarantees the ego and target footprints never intersect.
\begin{equation}
R_{ego}(t)\cap R_{target}(t)=\emptyset \quad \forall t\in[0,T_F],
\tag{CA}
\label{eqn:nocollisions_constraint}
\end{equation}
\noindent\textbf{Dynamic feasibility (DF):} constrains the ego’s acceleration to the tire–road grip limits, where $a_{\mathrm{lat}}(t)$ and $a_{\mathrm{lon}}(t)$ are the lateral/longitudinal components of the trajectory acceleration $\ddot{\mathcal{T}}_{ego}(t)$, and $v(t)=|\dot{\mathcal{T}}_{ego}(t)|$.
\begin{equation}
\Bigg(\frac{a_{\mathrm{lat}}(t)}{a_{\max,\mathrm{lat}}(v(t))}\Bigg)^{2} +
\Bigg(\frac{a_{\mathrm{lon}}(t)- c_{\mathrm{lon}}(v(t))}{a_{\max,\mathrm{lon}}(v(t))}\Bigg)^{2} \leq 1,
\quad \forall t
\tag{DF}
\label{eqn:dynamics_ellipse_constraint}
\end{equation}
\noindent\textbf{Finish-ahead (FA):} requires the ego to end at least $\Delta s_F$ further along the ORL than the target by the end of the planning horizon. 
\begin{equation}
s_F^{ego}-s_F^{target}>\Delta s_F,\qquad
s_F^{ego}=\pi_{ORL}(\mathcal{T}_{ego}(T_F)),
\tag{FA}
\label{eqn:finish_ahead_constraint}
\end{equation}
where $s_F^{ego} = \pi_{ORL}(\mathcal{T}_{ego}(T_F))$ is projection of the ego position at $t=T_{F}$ and $s_F^{target}$ is the corresponding target vehicle projection.

\noindent\textbf{Endpoint rejoin (E):} requires the ego trajectory to merge back on the ORL with matching velocity.  
\begin{equation}
    \mathcal{T}_{ego}(T_F)\in\mathbb{ORL}, \quad
    \dot{\mathcal{T}}_{ego}(T_F)=\mathbf{v}_{ORL}(s_F^{ego}).
    \tag{E}
      \label{eqn:endpoint}
\end{equation}

Together, \eqref{eqn:initial_pos}, \eqref{eqn:ontrack_constraint}, \eqref{eqn:nocollisions_constraint}, \eqref{eqn:dynamics_ellipse_constraint}, \eqref{eqn:finish_ahead_constraint}  and \eqref{eqn:endpoint} define the mathematical requirements of a valid overtaking maneuver. 
Our method enforces these constraints using composite B'ezier curves (CBCs), whose properties make them a natural parameterization for feasible racing trajectories.

\noindent \textbf{Assumptions:} 
We make the following assumptions, consistent with prior work on autonomous racing.
(1) Track boundaries and the ORL are known a priori from offline computation (standard in racing).
(2) A target-trajectory predictor is available (e.g., \cite{barte,mtr++}); our planner is predictor-agnostic.
(3) The target footprint is a bounding polygon aligned with the predictor’s velocity at each time step; i.e. target body slip is assumed negligible (0) unless estimated, though the framework can incorporate nonzero slip. Collision checks use a GLR-style evaluation \cite{GLR}.
(4) Ego acceleration limits $a_{\max,\mathrm{lat}}(\cdot),a_{\max,\mathrm{lon}}(\cdot)$ come from tire dynamics/Pacejka-type models (typical in model-based control).
(5) Ego state estimation is treated as known (we do not model uncertainty).
(6) While DBF-MA extends to multiple opponents, this paper lists the constraints for a single target vehicle for clarity.

\section{Background}
\label{sec:background}

We represent trajectories $\mathcal{T}(t)$ using Composite B\'ezier Curves (CBCs). A CBC is a piecewise polynomial spline in the Bernstein basis~\cite{bcurves_book_patrikalakis}, composed of multiple B\'ezier segments connected to form a longer curve. 
We briefly review the definition of B\'ezier curves and the key continuity properties of CBCs that are central to our method.
A B\'ezier curve (in $\mathbb{R}^2$) is defined as a convex combination of Bernstein polynomials. 
For $k+1$ control points ${\mathbf{C}0, \mathbf{C}1, \dots, \mathbf{C}k}\subset\mathbb{R}^2$, the curve $\mathbf{P}:[0,1]\to\mathbb{R}^2$ is given by
\begin{equation}
b_{i,k}(u) = \binom{k}{i}(1-u)^{k-i}u^i,
\label{eqn:bernstein_polynomials}
\end{equation}
\begin{equation}
\mathbf{P}(u) = \sum_{i=0}^k b_{i,k}(u)\mathbf{C}_i
\label{eqn:bezier_curve}
\end{equation}

\subsection{Composite B\`ezier Curves}
A Composite B\'ezier Curve is a spline consisting of $N_S$ connected B\'ezier segments of order $k$. 
A trajectory $\mathcal{T}(t)$ can be parameterized as:
\begin{equation}
u(t) = \frac{t - t_j}{t_{j+1}-t_j}, \quad
\mathcal{T}(t) = \mathbf{P}_j(u(t)),
\end{equation}
where $\mathbf{P}j$ denotes the $j$-th segment defined over $[t_j,t{j+1}]$. 
In this work, segments are uniformly spaced in time, $t{j+1}-t_j = T_F/N_S ; \forall j$. Non-uniform spacing is left to future work.

We denote control point $j$ of segment $i$ as $\mathbf{C}_{i,j}$. First-order continuity between consecutive segments is enforced ~\cite{bcurves_book_patrikalakis} via
\begin{equation}
    \mathbf{C}_{i,k} = \mathbf{C}_{i+1,0}\text{ } \forall i<N_S-1
    \label{eqn:cbc_first_order_continuity}
\end{equation}
\begin{equation}
    \mathbf{C}_{i,k} - \mathbf{C}_{i,k-1} = \mathbf{C}_{i+1,1} - \mathbf{C}_{i+1,0}\text{ } \forall i<N-1
    \label{eqn:cbc_second_order_continuity}
\end{equation}

Thus a CBC is fully defined by its control points and naturally provides a $C^1$ continuous parameterization. 
We adopt CBCs as the representation of local overtaking trajectories $\mathcal{T}{ego}(t)$. 
Figure~\ref{fig:t_theta} illustrates an example curve with its control points $\mathbf{C}{i,j}$.
We next present the Multi-Agent Differential Bayesian Filtering (DBF-MA) algorithm, which leverages CBCs as the underlying trajectory representation.


\section{Multi-Agent Differential Bayesian Filtering}
\label{sec:methodology}
Our methodology builds on the Differential Bayesian Filtering (DBF) framework introduced in~\cite{ral_dbf}. 
In DBF, trajectory synthesis is cast as Bayesian inference over a low-dimensional parameterization $\theta$ of trajectories $\mathcal{T}_\theta$. The key elements are:
\begin{enumerate}
\item A parameter vector $\theta$ defines a family of trajectories with prior distribution $p(\theta)$.
\item A likelihood function $p(x\mid\theta)$ encodes how well $\mathcal{T}_\theta$ satisfies continuity, track-keeping, and dynamic feasibility constraints.
\end{enumerate}
Sequential Monte Carlo (SMC) methods are then used to approximate the posterior distribution
\begin{equation}
p(\theta \mid x) = \frac{p(x\mid \theta),p(\theta)}{\int p(x\mid \theta),p(\theta),d\theta},
\label{eqn:bayes_generic}
\end{equation}
where $x$ represents evidence in the form of optimality constraints. In~\cite{ral_dbf}, the likelihood was limited to single-agent racing and encoded only vehicle dynamics and track boundaries.

We extend DBF to the multi-agent overtaking setting by introducing a likelihood that incorporates overtaking constraints. 
For clarity, we use $p(\theta)$ and $p(\mathcal{T}_\theta)$ interchangeably: $p(\theta)$ is a prior distribution over trajectories, and $p(\theta \mid x)$ is the posterior distribution of feasible overtaking trajectories.

DBF-MA now has two main components:
\begin{enumerate}
    \item A trajectory parameterization $\mathcal{T}_\theta$ that specifies how $\theta$ generates a continuous curve.
    \item A likelihood function $p(x\mid \theta)$ that encodes the probability that $\mathcal{T}_\theta$ satisfies all overtaking constraints.
\end{enumerate}

The following subsections describe these two pieces. 
First, we introduce our parameterization of $\mathcal{T}_\theta$ using composite CBCs. Then we detail our likelihood construction.

\subsection{Trajectory Parameterization with CBCs}
\label{subsec:t_theta_specification}
We represent an overtaking trajectory $\mathcal{T}_\theta$ as a composite B'ezier curve (CBC) with $N_S$ cubic segments ($k=3$). Higher-order curves were explored, but found to increase computational cost without improving performance.
The free parameters of our Bayesian model are the control points $\mathbf{C}_{i,j}; \mid i\in[0,1,...N_S-1],\text{ }j\leq 3$ of the CBC, with specific constraints to guarantee initial and terminal conditions.

\noindent\textbf{Satisfying the continuity constraint \eqref{eqn:initial_pos}}:
To satisfy \eqref{eqn:initial_pos}, we fix the first two control points of the first segment:
\begin{equation}
\mathbf{C}_{0,0} = \mathbf{x}_0, \qquad
\mathbf{C}_{0,1} = \mathbf{x}_0 + \tfrac{T_F}{N_S}\mathbf{v}_0,
\label{eqn:Ttheta_cp0}
\end{equation}
which ensures
\begin{equation}
\mathcal{T}_\theta(0) = \mathbf{x}_0, \qquad
\dot{\mathcal{T}}_\theta(0) = \mathbf{v}_0.
\label{eqn:Ttheta_init}
\end{equation}
meaning that $\mathcal{T}_\theta$ satisfies \eqref{eqn:initial_pos} $\forall \theta$.

\noindent \textbf{Satisfying the endpoint constraint \eqref{eqn:endpoint}:}
We fix the last two control points of the final segment of the CBC
by letting $s_F^{ego}$ (the Frenet coordinate of the ego's terminal position along the ORL) be the final free parameter in $\theta$. 
\begin{equation}
\mathbf{x}_{end} = \mathbf{p}_{ORL}(s_F^{ego}), \qquad
\mathbf{v}_{end} = \mathbf{v}_{ORL}(s_F^{ego}),
\end{equation}
\begin{equation}
\mathbf{C}_{N_S-1,k}   = \mathbf{x}_{end}, \
\mathbf{C}_{N_S-1,k-1} = \mathbf{x}_{end}-(T_F-t_{N_S-1})\mathbf{v}_{end},
\label{eqn:Ttheta_end_cps}
\end{equation}
which guarantees
\begin{equation}
    \mathcal{T}_\theta(T_F) = \mathbf{x}_{end} \qquad
    \dot{\mathcal{T}}_\theta(T_F) = \mathbf{v}_{end}
    \label{eqn:satisfy_final_vel}
\end{equation}
and $\mathcal{T}_\theta$ therefore satisfies \eqref{eqn:endpoint} $\forall \theta$.

All intermediate CBC segments are connected with $C^1$ continuity enforced via \eqref{eqn:cbc_first_order_continuity} and \eqref{eqn:cbc_second_order_continuity}. 
Our parameters, $\theta$, are therefore defined as:
\begin{equation}
    \theta=[
    \mathbf{C}_{0,2}, \mathbf{C}_{0,3}, ...  \mathbf{C}_{N_S-2,2}, \mathbf{C}_{N_S-2,3}, s_F^{ego}
    ]
    \label{eqn:full_theta_vector}
\end{equation}

\begin{figure}
    \centering
    \includegraphics[width=0.9\columnwidth]{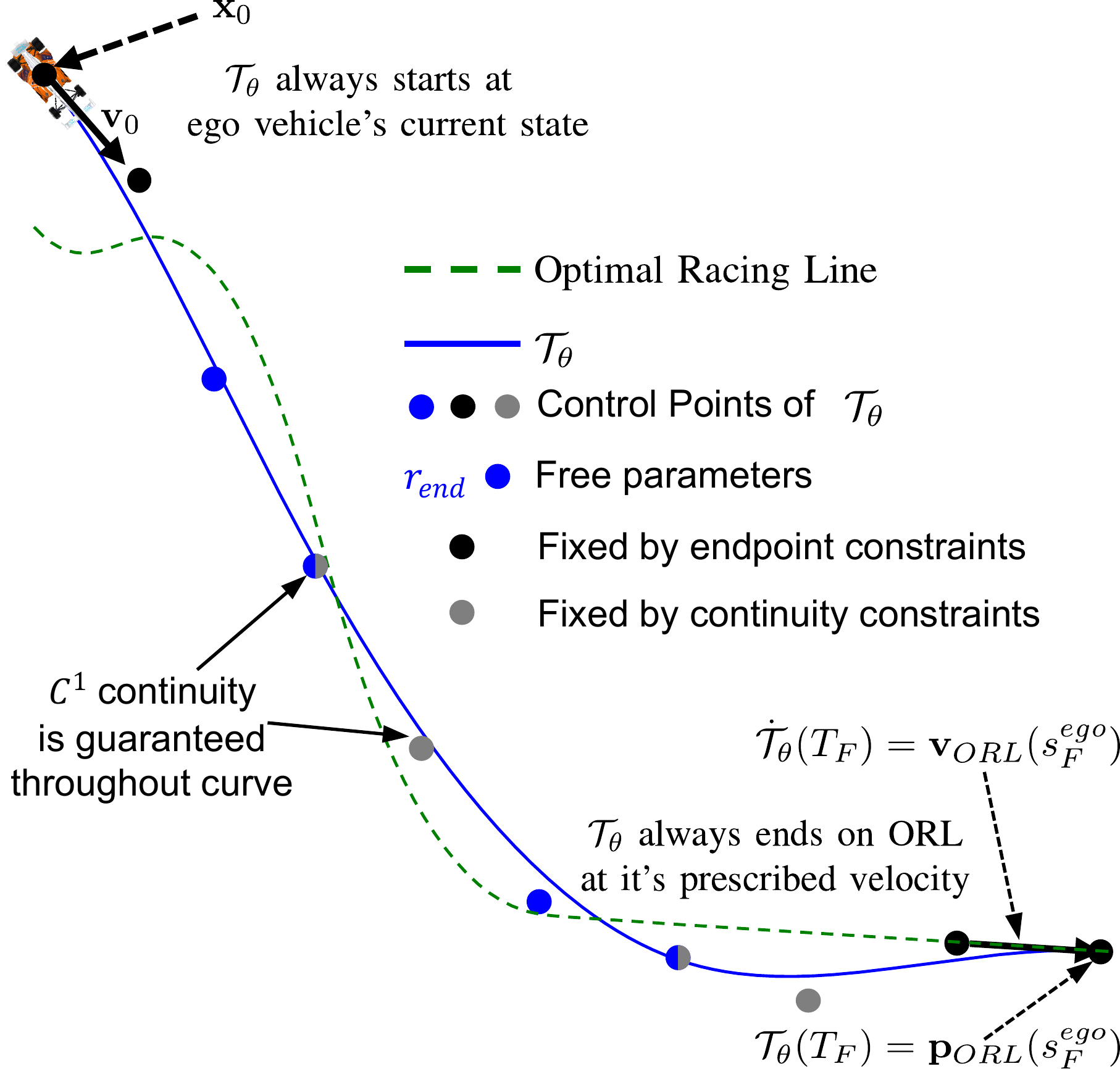}
    \caption{Example ($N_S=3$) of a CBC trajectory $\mathcal{T}_\theta$. Black control points are fixed to satisfy initial and endpoint constraints; blue points are free parameters in $\theta$. Our approach guarantees that $\mathcal{T}_\theta$ will satisfy \eqref{eqn:initial_pos} and \eqref{eqn:endpoint}}
    \label{fig:t_theta}
\end{figure}
An example of such a $\mathcal{T}_\theta$ is shown in Figure \ref{fig:t_theta}. 
The various components that define the constrained control points of the CBC are marked in black.  The control points left as free parameters are marked with blue dots.  The endpoint constraints at, $\mathcal{T}_{\theta}(T_F)$, follow from the ORL at $s_F^{ego}$. 
This parameterization offers three key benefits:
\begin{enumerate}
    \item $\mathcal{T}_\theta$ always satisfies \eqref{eqn:initial_pos} and \eqref{eqn:endpoint} for any $\theta$.
    \item The trajectory is low-dimensional yet expressive: $4(N_S-1)+1$ parameters make SMC inference tractable.
    \item $C^1$ continuity is guaranteed, yielding smooth entry into and exit from the overtaking maneuver.
\end{enumerate}
We now use this parameterization to construct the likelihood function $p(x\mid \theta)$.

\subsection{Likelihood Function $p(\theta \mid x)$}
Since \eqref{eqn:initial_pos} and \eqref{eqn:endpoint} hold for any $\mathcal{T}_\theta$ by construction, the likelihood models the remaining constraints: collision avoidance \eqref{eqn:nocollisions_constraint}, track-keeping \eqref{eqn:ontrack_constraint}, and dynamic feasibility \eqref{eqn:dynamics_ellipse_constraint}. 
The finish-ahead condition \eqref{eqn:finish_ahead_constraint} is enforced during SMC; (see Sec.~\ref{subsec:smc_procedure}).
We adopt a separable form:
\begin{equation}
p(x\mid\theta) = p_{(CA)}(\theta),p_{(TK)}(\theta),p_{(DF)}(\theta),
\label{eqn:full_likelihood}
\end{equation}
where each factor is a time-aggregated ``no-violation" probability derived from the log-odds integral of an instantaneous risk measure.


\noindent\textbf{Collision Avoidance Likelihood $p_{(CA)}(\theta)$}:
Let
\begin{equation}
    L_{(CA)}(t,\theta) = Pr[R_{ego}(t)\cap R_{target}(t) \neq \emptyset \vert \mathcal{T}_\theta]
    \label{eqn:instantaneous_boundary_violation_prob}
\end{equation}
be the instantaneous collision probability that can be computed with the Gauss--Legendre Rectangle (GLR) method~\cite{GLR}, which represents collision events with a Poisson counting process.
Here $Pr[R_{ego}(t)\cap R_{target}(t) \neq \emptyset \vert \mathcal{T}_\theta]$ is the instantaneous probability of collision at $t$.
Modeling collisions as a Poisson process with time-varying hazard given by the odds $\frac{L_{(CA)}}{1-L_{(CA)}}$, the probability of \emph{no collision} over $[0,T_F]$ is:
\begin{equation}
    p_{(CA)}(\theta) = 1.0-\exp\Bigl(-\int_0^{T_F}\frac{ L_{(CA)}(t,\theta)}{1.0- L_{(CA)}(t,\theta)}dt\Bigr )
    \label{eqn:collision_prob_glr}
\end{equation}
We refer the interested reader to ~\cite{GLR} for a detailed explanation of the GLR algorithm and its implementation. 
This is calculating, for a given opponent prediction, and a candidate ego trajectory, the probability that there is no collision based on rectangular polygon overlap at any point in time.

\begin{figure}
    \centering
    \includegraphics[width=0.9\linewidth]{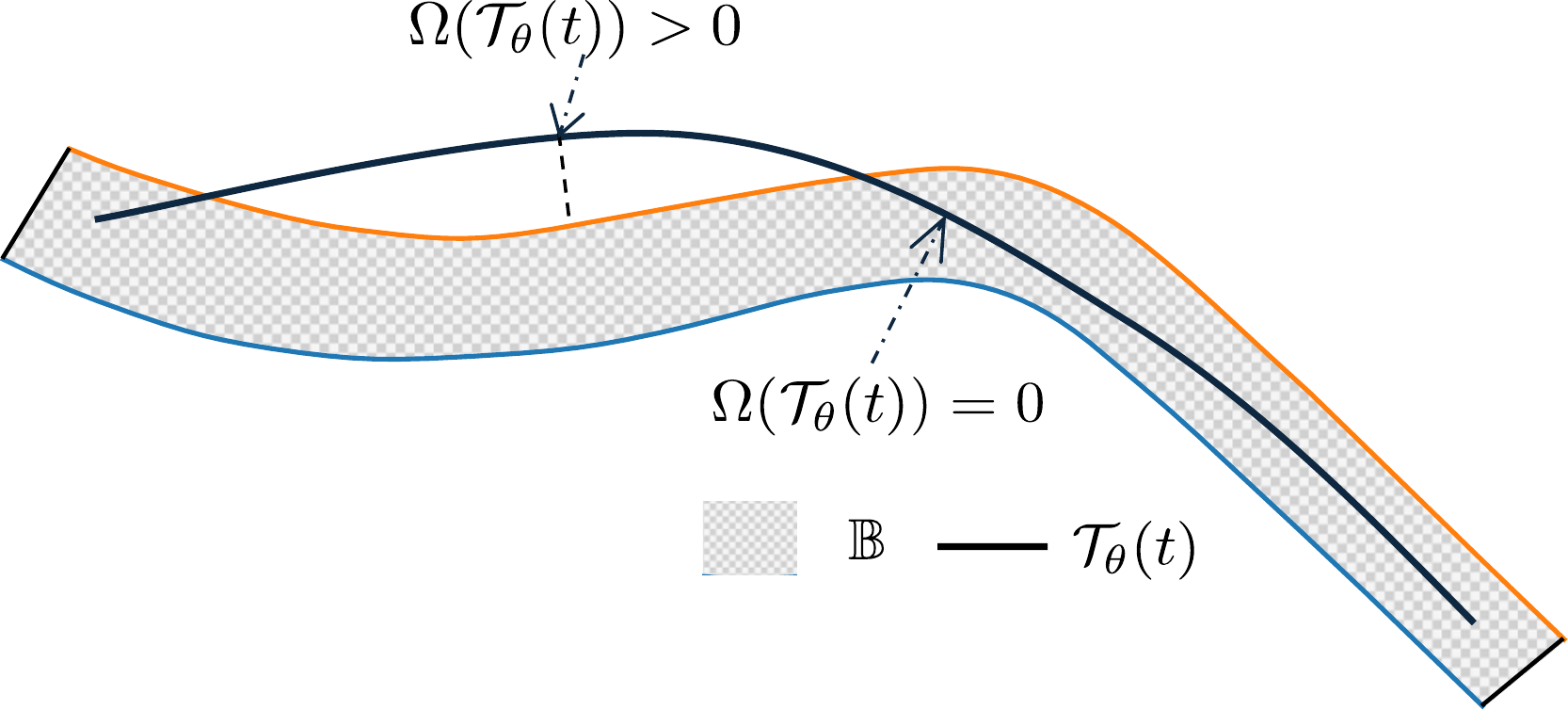}
    \caption{An illustration of $\Omega(\mathcal{T}_\theta(t))$}
    \label{fig:Omega_T_illustration}
\end{figure}

\noindent\textbf{Track Keeping Likelihood,} $p_{(TK)}(\theta)$:
Let $\Omega(\mathcal{T}_\theta(t))$ denote the (unsigned) distance from $\mathcal{T}_\theta(t)$ to the drivable set $\mathbb{B}$, so $\Omega=0$ iff $\mathcal{T}_\theta(t)\in\mathbb{B}$. 
We map this distance to an instantaneous boundary-violation probability via
\begin{equation}
    L_{(TK)}(t,\theta) = \Phi\Bigr(\frac{\Omega(\mathcal{T}_\theta(t))}{\sigma_{B}}\Bigl) - 0.5
    \label{eqn:instantaneous_boundary_violation_prob}
\end{equation}
where $\Phi(\cdot)$ is the standard normal CDF and $\sigma_B$ is a scale hyperparameter. 
Thus $L_{(TK)}=0$ when the trajectory is on-track ($\mathcal{T}_\theta(t)\notin \mathbb{B}$) and $L_{(TK)}\rightarrow 1$ with increasing out-of-bounds distance, with $\sigma_L$ controlling how quickly $L_{(TK)}(t)$ approaches $1$. Figure \ref{fig:Omega_T_illustration} depicts an example trajectory annotated with an area where $\mathcal{T}_\theta(t) \in \mathbb{B}$, and therefore $\Omega(\mathcal{T}(t)\bigr)=0$, as well as an area where $\mathcal{T}_\theta(t) \notin \mathbb{B}$, and therefore $\Omega(\mathcal{T}(t)\bigr)>0$.

Our overall probability that $\mathcal{T}_\theta$ always stays within bounds over the horizon takes a similar integral log-odds form:
\begin{equation}
    p_{(TK)}(\theta) = 1.0-\exp\Bigl(-\int_0^{T_F}\frac{L_{(TK)}(t)}{1-L_{(TK)}(t)}dt\Bigr)
    \label{eqn:overall_inbounds_prob}
\end{equation}
$p_{(TK)}(\theta)$ is, therefore, the probability of satisfying the track keeping constraint ~\eqref{eqn:ontrack_constraint}. 

\noindent\textbf{Dynamics Feasibility Likelihood,} $p_{(DF)}(\theta)$:
The dynamic feasibility constraint \eqref{eqn:dynamics_ellipse_constraint} is equivalent to the well-known
\emph{traction ellipse} (or GG-diagram), which defines the feasible region of
$(a_{\mathrm{lon}},a_{\mathrm{lat}})$ at a given velocity. 
Its shape depends on tire slip and friction limits, often modeled with Pacejka tire dynamics. 
For convenience, we denote this feasible set at $v(t)=|\dot{\mathcal{T}}_\theta(t)|$ as $\mathbb{T}(v(t))$.

\begin{figure}
    \centering
    \includegraphics[width=0.90\linewidth]{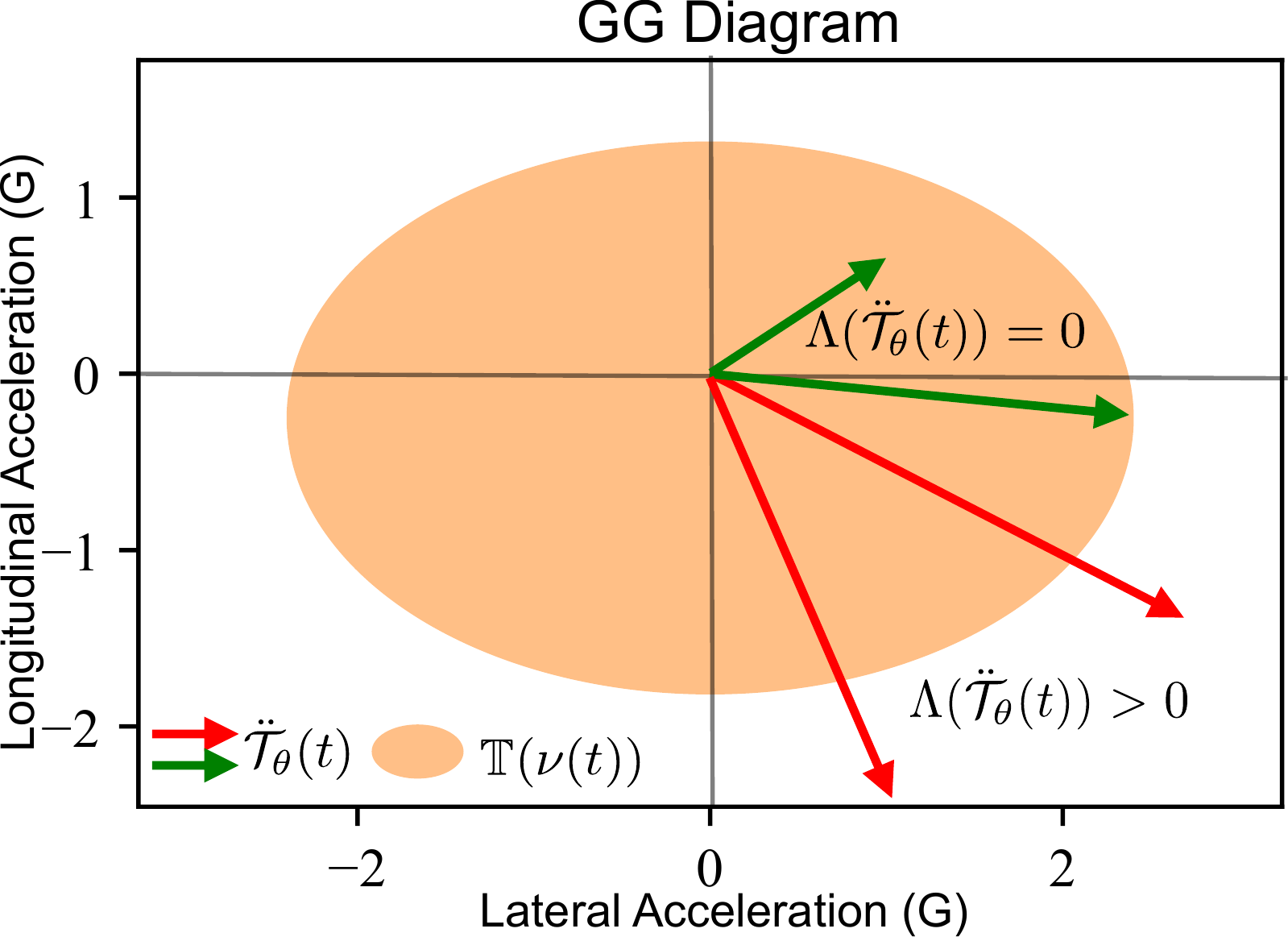}
    \caption{A GG diagram showing $\mathbb{T}$ and $\Lambda$. The green acceleration vectors have a $\Lambda$ value of 0, while the red vectors have a strictly positive $\Lambda$.}
    \label{fig:lambda_accel}
\end{figure}

Define $\Lambda(\ddot{\mathcal{T}}_\theta(t))$ as the minimum (unsigned) radial offset from $\ddot{\mathcal{T}}_\theta(t)$ to the ellipse boundary in the $a_{\mathrm{lat}}$--$a_{\mathrm{lon}}$ (GG) plane, so $\Lambda=0$ if feasible and $\Lambda>0$ otherwise. 
This is illustrated in figure \ref{fig:lambda_accel} where $\mathbb{T}$, is shown in orange, with two acceleration vectors that fall inside of $\mathbb{T}$ shown in green. Two acceleration vectors that fall outside of $\mathbb{T}$ (and therefore with  $\Lambda > 0$) are shown in red.
We set:
\begin{equation}
    L_{(DF)}(t,\theta) = \Phi\Bigl(\frac{\Lambda(\ddot{\mathcal{T}}_\theta(t)\bigr)}{\sigma_{D}}\Bigr) - 0.5
    \label{eqn:instantaneous_dynamics_violation_prob}
\end{equation}
with scale $\sigma_D$, and aggregate as
\begin{equation}
    p_{(DF)}(\theta) = 1.0-\exp\Bigl(-\int_0^{T_F}\frac{L_{(DF)}(t)}{1-L_{(DF)}(t)}dt\Bigr )
    \label{eqn:overall_dynamics_violation_prob}
\end{equation}
$p_{(DF)}(\theta)$ is the probability that $\mathcal{T}_\theta$ satisfies the dynamic feasibility constraint \eqref{eqn:dynamics_ellipse_constraint}.


\noindent\textbf{Overall Likelihood Function}:
As stated in Eq.~\eqref{eqn:full_likelihood}, the overall likelihood $p(\theta|x)$ is the joint probability that $\mathcal{T}_\theta$ simultaneously satisfies collision avoidance (CA), track keeping (TK), and dynamic feasibility (DF). 
This formulation enables a sampling-based sequential Monte Carlo procedure to infer trajectories that remain collision-free, within track limits, and dynamically feasible, and enforce the last remaining constraint of finishing ahead of the opponent.

\subsection{SMC Bayesian Inference for Autonomous Overtaking}
\label{subsec:smc_procedure}

\begin{figure}
    \centering
    \includegraphics[width=0.95\linewidth]{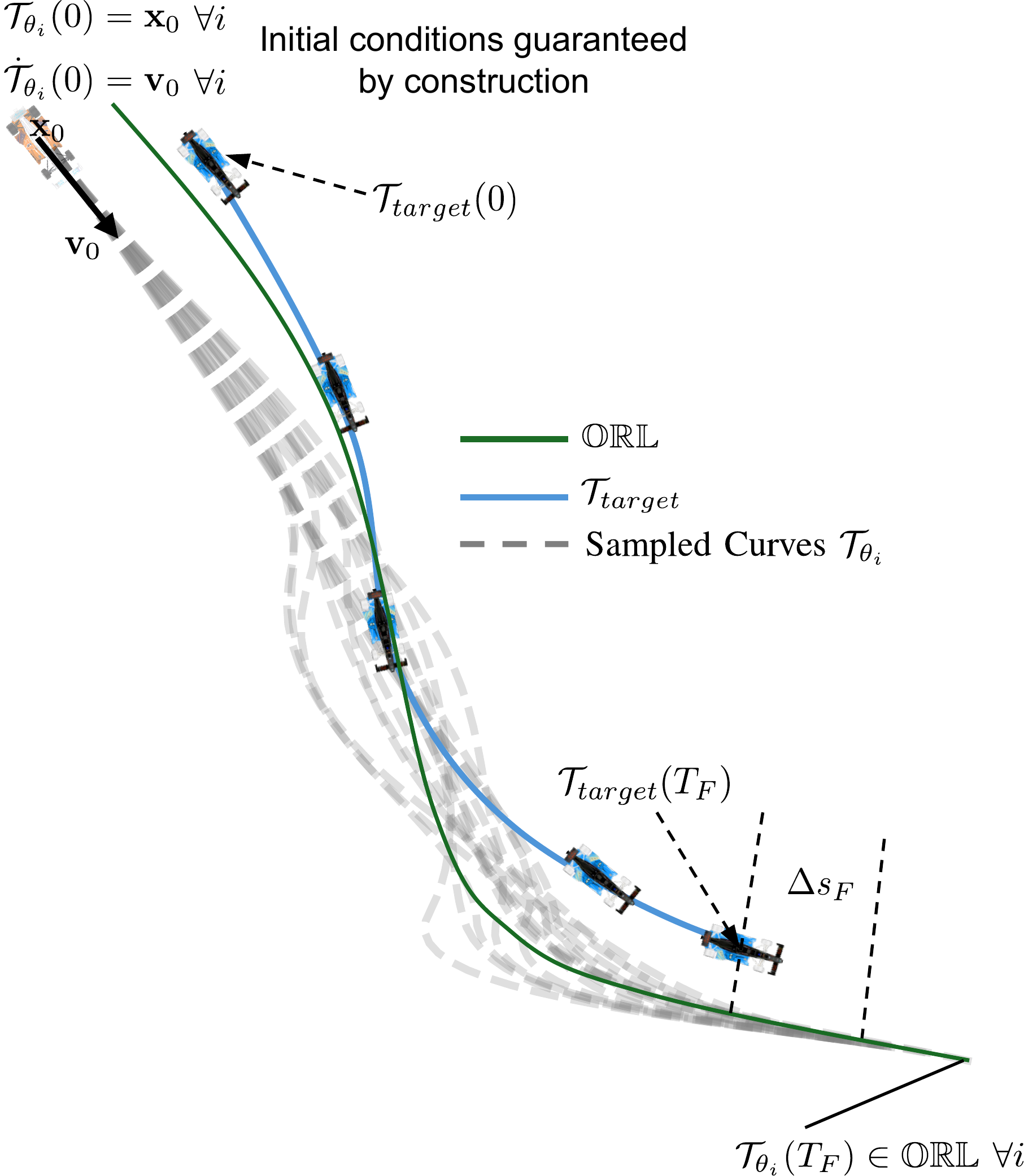}
    \caption{One step of the DBF-MA algorithm. Each grey curve (particle) is assigned a probability defined by $p(\theta_i \vert x)$. The curves are resampled according to these probabilities in the subsequent step of the algorithm.}
    \label{fig:biao_illustration}
\end{figure}

Given our likelihood function $p(\theta|x)$, we employ Sequential Monte Carlo (SMC) inference to draw samples from the posterior distribution of feasible overtaking trajectories. 
This choice is natural since $p(\theta|x)$ is typically non-convex, and SMC provides a principled way to approximate such distributions via resampling. 
We view the problem as a particle filter over the space of trajectories, where each particle corresponds to a parameter vector $\theta$ defining a CBC overtaking trajectory.
We approximate the distribution $p(\theta)$ using $N_P$ particles, ${\theta_1, \theta_2, \ldots, \theta_{N_P}}$, each representing a candidate trajectory. 
The inference proceeds as follows:
\begin{enumerate}
\item \textbf{Initialization:} Set all $N_P$ particles to a least-squares fit of the ORL over the prediction horizon, subject to the ego vehicle's current state. This initializes the prior distribution $p(\theta)$ with the ORL as its mean trajectory.
\item \textbf{Perturbation:} Add Gaussian noise with zero mean and covariance $\Sigma_\theta$ to each parameter vector, generating candidate trajectories.
\item \textbf{Endpoint clipping:} Enforce the finish-ahead constraint by clipping each $s_{F,i}^{ego}$ such that the finish ahead (FA) constraint $s_{F,i}^{ego} \geq s_F^{target} + \Delta s_F$  is enforced. This guarantees that all particles terminate ahead of the target vehicle on the ORL, based on the prediction for the target vehicle. 
\item \textbf{Scoring:} Evaluate each particle with the likelihood $p(\theta_i|x)$, producing a weight $P_i$ that reflects its probability of satisfying the overtaking constraints.
\item \textbf{Resampling:} Resample $N_P$ particles according to their weights, reinforcing trajectories with higher likelihood.
\item \textbf{Stopping:} Repeat steps 2–5 (up to a maximum number of iterations, $N_{iter}$) until at least one particle achieves likelihood $p(\theta_i|x) \geq 1-\epsilon$, where $\epsilon$ specifies the maximum tolerable risk. This yields an overtaking trajectory that satisfies all constraints with high confidence. If no such particle is found after $N_{iter}$ iterations, overtaking is presumed impossible in the current planning context.
\end{enumerate}
Figure~\ref{fig:biao_illustration} illustrates a single resampling step of this procedure: each grey curve is assigned a likelihood score, and particles are resampled according to these scores. The complete algorithm is summarized in Algorithm~\ref{alg:dbf}.

\begin{algorithm}
\SetAlgoLined
    \setlength{\textfloatsep}{0pt}
      //\text{Initialize all particles with local slice of the ORL}\\
      $\{\theta_1, \theta_2, ... \theta_{N_P}\} \gets \text{LSQ fit of ORL } s.t.$ \eqref{eqn:initial_pos}\\
      
    \For{0...$N_{iter}-1$}{
            
        $p_{1,i} \gets p_{(CA)}(\theta_i)$ //$Pr\{\text{Collision-free}\}$
        $p_{2,i} \gets p_{(TK)}(\theta_i)$ //$Pr\{\text{Within track limits}\}$
        $p_{3,i} \gets p_{(DF)}(\theta_i)$ //$\{\text{Dynamically feasible}\}$ 
        %
        $P_i \gets p_{1,i}p_{2,i}p_{3,i}$ //$Pr\{\mathcal{T}_{\theta_i}$\text{ meets all criteria}\}

        $I \gets \underset{i}{argmax}\text{ }P_i$ //{Index of best particle}

        \If { $P_{I}\leq 1-\epsilon$ }{
            //\text{Best particle has low enough risk}
            //\text{return its associated trajectory}
            
            \textbf{return } $\mathcal{T_{\theta_{I}}}$
        }
        //\text{Resample particles according to }$P_i$
        $\{\theta_1, \theta_2, ... \theta_{N_P}\} \sim P_i$

        //\text{Add gaussian noise to each resampled particle}
        $\theta_i \gets \theta_i + \mathcal{N}(0, \Sigma_\theta)$
        
        $s_{F,i}^{ego} \gets max(s_{F,i}^{ego}, s_{F}^{target} + \Delta s_F)$
        
    }
    
    //\text{Max iterations reached and no solution found}\\
    \textbf{return }\text{ Overtaking is currently impossible}
    \caption{DBF-MA}
    \label{alg:dbf}
\end{algorithm}

\section{Experimental Results}
\label{sec:results}
\subsection{Experimental Setup}
\label{subsec:experimental_setup}
We evaluate DBF-MA against two baseline methods: Predictive Spliner\cite{predictive_spliner} and the Graph-Based Planner (GP)\cite{graph_planner_og}. 
Evaluation is performed in closed-loop simulation over 1800 overtaking scenarios generated across three Formula 1 tracks (Australia, Monza, Britain).  
Simulations are run in \textit{Cavsim}, the high-fidelity physics simulator used by our autonomous racing team in the Indy Autonomous Challenge. 
Cavsim provides agent-level dynamics, tire models, and track geometry within a ROS2-based environment.
A scenario is constructed as follows:
\begin{enumerate}[noitemsep]
    \item The ego vehicle is initialized at a randomly sampled location on the ORL.
    \item The target vehicle is placed 0.5~s ahead on the same ORL.
    \item The target vehicle follows the ORL at a reduced speed.  
\end{enumerate}
We evaluate 600 scenarios each with the target constrained to $88\%$, $76\%$, or $64\%$ of the ORL speed profile, which has speeds up to 150 MPH. 
The ego vehicle is unrestricted.  Scenarios are distributed uniformly across the three tracks: 200 scenarios per target speed on each track.

The target vehicle is constrained to the ORL to reflect consistency with real racing rules used in current autonomous racing competitions, where reactive blocking or weaving is explicitly prohibited (e.g. IAC, and A2RL). This choice also allows us to isolate overtaking as a planning problem, without confounding effects of defensive maneuvers. 
Extensions to actively defending opponents will be addressed in future work.


\noindent \textbf{Closed-loop simulation:} At scenario start, both vehicles follow the ORL. 
The planner under test attempts to generate an overtaking trajectory; if successful, the ego vehicle transitions from the ORL to the planned trajectory, re-planning as needed. Simulations terminate if an overtake is not completed within 80~s.  
All planners are provided with ground-truth target trajectories to isolate planning performance from prediction errors. A common model-based predictive controller (MPC) is used as the low-level controller across methods to ensure consistent trajectory following.  

\noindent \textbf{Dynamic feasibility constraints:}
The mapping from speed to the friction ellipse, $\mathbb{T}(v(t))$, is a linear model consistent with the typical performance of an IndyNXT vehicle.  Longitudinal limits linearly increase from $1.5$G at rest to $0$ at a speed of $165$ MPH, braking limits from $-1.5G$ to $-2.5$G, and lateral acceleration limits scale from $\sim2G$ to $\sim3.5G$ over the same speed range. Figure \ref{fig:Omega_T_illustration} shows $\mathbb{T}(t)$ at $v(t)=30$ MPH.

\noindent \textbf{Hyperparameters} for DBF-MA are listed in Table~\ref{table:hyperparams}. $T_F$ was selected as the duration of a typical overtaking maneuver, $\Delta s_F$ was chosen as 3 car lengths, and the remaining hyperparameters were tuned manually. Baseline parameters follow their respective publications. All experiments are run on an Intel Xeon Silver 4210 CPU with an NVIDIA Quadro RTX 4000 GPU for all the methods.

\noindent \textbf{Metrics:}  
We report the following performance metrics:  
\begin{enumerate}
    \item \textbf{Success Count:} Number of successful overtakes without any constraint violations. Higher is better. 
    \item \textbf{Dynamics Violation Severity (DVS):} Euclidean norm between the ego vehicle's acceleration and $\mathbb{T}(t)$. Larger values indicate a tendency to violate dynamic limits.
    \item \textbf{Collisions:} Number of collisions between ego and target.  
    \item \textbf{Cross-Track Error (CTE):} Average Euclidean deviation between ego position and its planned trajectory; large CTE indicates infeasible plans.  
    \item \textbf{Time to Overtake (TTO):} Duration required to complete an overtake.  
\end{enumerate}

\begin{table}
    \centering
    \resizebox{\columnwidth}{!}{
    \begin{tabular}{|l|c|c|c|c|c|c|c|c|}
        \hline
        Hyperparameter & $N_{iter}$ & $N_P$ & $N_S$ & $\sigma_D$ ($\unit{\meter\per\square\second}$) & $\sigma_B$ ($\unit{\meter}$) & $\Sigma_\theta$ ($\unit{\meter}$)  & $\Delta s_F$ ($\unit{\meter}$) & $T_F$ $(\unit{\second})$   \\ \hline 
        Value & 8 &    256 & 2       & 0.2    & 0.75  & 0.875$\mathbf{I}$   & 15.6 & 8 \\ \hline 
        
    \end{tabular}
    }
    \caption{Hyperparameter values for DBF-MA in our experiments}
    \label{table:hyperparams}
\end{table}

\subsection{Simulation Results}
\begin{table}[!h]
    \centering
    \resizebox{0.975\columnwidth}{!}{%
    \begin{tabular}{|l|c|c|c|c|c|c|}
        \hline
        Method & Target Scale & Successes & \makecell{DVS (\unit{\metre\per\square\second})} & Collisions & CTE (\unit{\metre}) & TTO (\unit{\second}) \\ 
        \hline
        \multirow{3}{4em}{\centering DBF-MA (Ours)}
                                        & .64 & 525 & .0009 & 8 & .173 & 13.19 \\ \cline{2-7}
                                        & .76 & 499 & .0033 & 6 & .183 & 16.43 \\ \cline{2-7}
                                        & .88 & 542 & .0031 & 0 & .16 & 16.43 \\ \cline{2-7} 
                                        & All & \textbf{1566} & \textbf{.0024} & \textbf{14} & \textbf{0.17} & 16.76 \\ \cline{2-7} \hlineB{3}
                                        
        \hline
        \multirow{3}{4em}{\centering Graph Planner}
                                        & .64 & 17 & 1.106 & 20 & .41 & 11.80 \\ \cline{2-7}
                                        & .76 & 6 & .839 & 1 & .52 & 20.21\\ \cline{2-7}
                                        & .88 & 1 & .573 & 0 & .77 & 4.87 \\ \cline{2-7} 
                                        & All & 24 & .569 & 21 & 0.57 & \textbf{12.3} \\ \cline{2-7} \hlineB{3}
                                        
        \hline
        \multirow{3}{4em}{\centering Predictive Spliner}
                                        & .64 & 6 & .913 & 270 & .72 & 28.87 \\ \cline{2-7}
                                        & .76 & 10 & .934 & 150 & .70 & 14.61 \\ \cline{2-7}
                                        & .88 & 32 & .811 & 138 & .71 & 8.5 \\ \cline{2-7} 
                                        & All & 48 & .886 & 558 & .71 & 17.32 \\ \cline{2-7}\hlineB{3}
    \end{tabular}
    }
    \caption{Experimental results on our simulation testbed.}
    \label{table:mainresults}
\end{table}
The results of our closed-loop experiments are summarized in Table~\ref{table:mainresults}. 
The Target Scale column specifies the speed restriction applied to the target vehicle in each of the 600 scenarios (64\%, 76\%, or 88\% of the ORL speed). 
Rows labeled All aggregate results across all 1800 scenarios: averages are reported for DVS, CTE, and TTO metrics, while total counts are reported for successes and collisions. 
The best performance for each metric is highlighted in bold.

Overall, DBF-MA achieves the highest success rate by a wide margin, with 1566 successful overtakes out of 1800 scenarios (87\%). 
By contrast, Predictive Spliner succeeds in only 48 cases, and the graph-based planner (GP) in just 24. 
DBF-MA also attains the lowest average Dynamics Violation Severity (DVS) and Cross-Track Error (CTE), demonstrating that its planned trajectories are both dynamically feasible and easier for the MPC path-follower to track.

GP relies on constant-acceleration segments between graph layers, which severely limits its ability to exploit corners for overtaking. 
As a result, GP almost exclusively attempts passes on long straights (typically the start–finish straight). 
This conservative behavior yields few collisions (21 in total), but also very few overtakes (24), and virtually none in cornering scenarios. When GP does succeed, it completes the maneuver relatively quickly, giving it the lowest average TTO among the three methods.

Predictive Spliner represents the opposite extreme: its high CTE and DVS values reveal that it generates trajectories beyond the feasible operating envelope of the ego vehicle. 
Although the nominal paths often pass the target, the MPC cannot track them reliably, leading to frequent collisions (558 in total). 
This problem is compounded by the method inheriting ORL speed profile regardless of curvature, which produces dynamically infeasible cornering maneuvers.

In contrast, DBF-MA consistently balances feasibility with competitiveness. It overtakes successfully in 1566 scenarios, with a relatively quick TTO, while maintaining the lowest DVS and CTE across all target speed conditions. 
These results confirm that explicitly modeling dynamic feasibility and collision risk within a Bayesian trajectory inference framework produces plans that are both safer and easier for the low-level controller to execute. However, DBF-MA collides on 14 scenarios where the algorithm took longer than average and resulted in a trajectory based on outdated collision-avoidance information. This problem is compounded on the straights where computation delays result in larger error in target vehicle predictions.  Future work would include improving the algorithm's computation time or planning from expected future states of the ego vehicle ahead of time.

\begin{figure}
    \centering
    \includegraphics[width=0.975\linewidth]{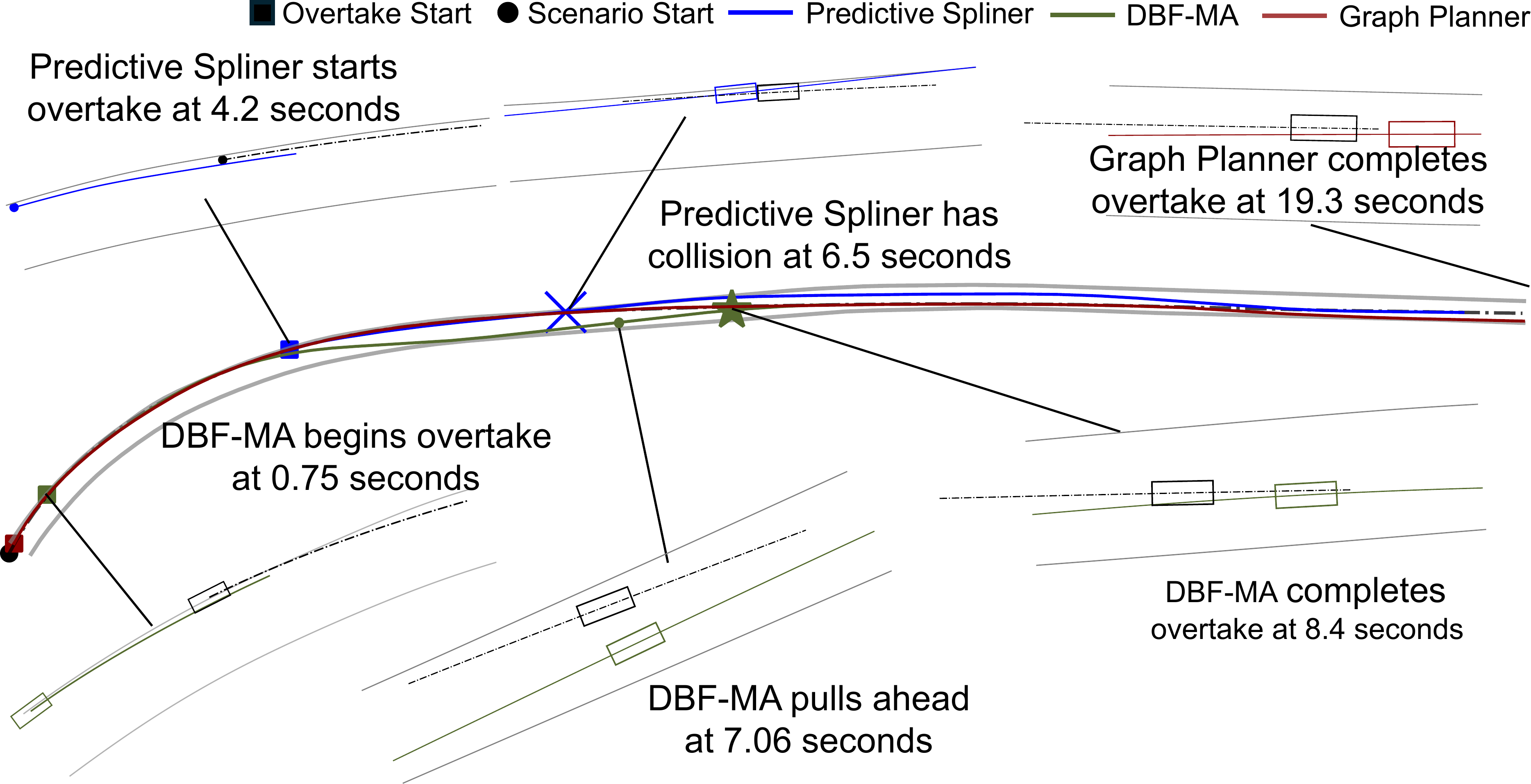}
    \caption{An example of each planner's closed-loop behavior.  Our method produces more feasible paths than Predictive Spliner and overtakes far more frequently than the graph planner.}
    \label{fig:example_outputs}
\end{figure}

Figure \ref{fig:example_outputs} shows an example of each method's behavior on a scenario from our experiments.  DBF-MA makes a clean overtake on the exit of a corner, completing the pass at 8.4 seconds after scenario start.  Predictive Spliner attempts an aggressive outside overtake and collides with the target vehicle at 4.2 seconds.  GP eventually overtakes the target, but requires 19.3 seconds to do so on the start-finish straight.

\subsection{Computational Analysis}

As in existing works~\cite{predictive_spliner,spliner}, we measure CPU/memory and computation time for each method on an Intel Xeon Silver 4210 CPU alongside an NVIDIA Quadro RTX 4000 GPU. These measurements are reported in table \ref{table:compresults}.  

The CPU\% values are total CPU utilization across all threads spawned by the process on multiple CPU cores, so values can exceed 100\%.  We also report CPU utilization per active thread. On a per-thread basis, all three methods have very similar CPU utilization, indicating the ROS2 infrastructure within the planner node's executable is accounting for much of the CPU usage.  

The computation time measurements measure only the core algorithm described in each method. Calls to ROS2 libraries were excluded.  On this metric, GP is the fastest algorithm, but this does not make up for its poor performance on other metrics.  DBF-MA is second-fastest, with an average planning time of 49.22$\unit{\milli\second}$, which is fast enough for good closed-loop performance.  DBF-MA also maps well to GPU acceleration given its highly parallel structure: all 3 of its likelihood functions for all particles can be computed simultaneously. Our PyTorch implementation of DBF-MA heavily utilizes the GPU, greatly accelerating its computation time. 

\begin{table}[!h]
    \centering
    \resizebox{\columnwidth}{!}{%
    \begin{tabular}{|l|c|c|c|c|c|}
        \hline
        Method & Computation Time (ms) $\pm \sigma$ & CPU\% $\pm \sigma$ & CPU\%/thread $\pm \sigma$ & Memory (GB)$\pm \sigma$ & Nthreads \\ \hline
        DBF-MA &                           49.22$\pm$17.77 & 99.62 $\pm$ 6.41 & 6.23 $\pm$ 0.32 & 1.57 $\pm$ 0.027 & 16 \\ \hline
        \makecell{Predictive Spliner} &   102.98$\pm$23.16 & 101.04 $\pm$ 5.1 & 6.31 $\pm$ 0.32 & 1.530 $\pm$ 0.022 & 16 \\ \hline
        Graph &                           8.16$\pm$ 9.64 & 69.23 $\pm$ 30.60 & 5.77 $\pm$ 2.55 & .161 $\pm$ 0.0005 & 12 \\ \hline
        
    \end{tabular}
    }
    \caption{Computational load metrics for each model.}
    \label{table:compresults}
\end{table}


%


\section{Conclusion and Future Work}
\label{sec:conclusion}

In conclusion, we present the DBF-MA algorithm: Differential Bayesian Filtering for Multi-Agent racing.  DBF-MA is a method of motion planning for overtaking scenarios in racing that combines a new way of specifying overtaking maneuvers using Composite B\'ezier Curves with sampling-based Monte Carlo Bayesian inference to synthesize collision-free and feasible overtaking motion plans.  DBF-MA offers continuity guarantees, both with the ego vehicles current state and with the end-state merging back to the ORL, that existing planners cannot.  We evaluate DBF-MA in a physics-based simulator, where it achieves a success rate of $84\%$ and outperforms existing motion planners on various racing metrics.  These results suggest DBF-MA is a promising approach for competitive autonomous racing, such as in the IAC or A2RL. Future work includes extending the approach to multiple target vehicle's in the planning horizon, exploring alternative methods of specifying a prior distribution of trajectories; such as a machine-learning or other data-driven technique; and integrating an interaction model into the framework where the ego trajectory and target trajectory are jointly distributed.
\bibliographystyle{IEEEtran}
\bibliography{references}

\end{document}